\documentclass[letterpaper, 10pt, conference]{ieeeconf}

\IEEEoverridecommandlockouts
\usepackage{graphicx}
\usepackage{amsmath}
\usepackage{amssymb}
\usepackage{booktabs}
\usepackage{multirow}
\usepackage{cite}
\usepackage{url}
\usepackage[table]{xcolor}
\usepackage{capt-of}
\usepackage{dblfloatfix}
\usepackage[hidelinks,breaklinks=true]{hyperref}

\graphicspath{{figures/}}

\newcommand{\method}{\textsc{PointCast}}

\title{\LARGE \bf
  \method: One World Model for Rigid, Articulated, and Deformable Object Manipulation
}

\author{Hantao Ye, Ross Worobel, Zhuoli Xie, Mingen Li, Houjian Yu,
  Youngjin Hong, and Changhyun Choi%
  \thanks{The authors are with the University of Minnesota, Minneapolis, MN
    55455, USA.
    {\tt\small \{ye000310, worob006, xie00507, li002852, yu000487, hong0745,
  cchoi\}@umn.edu}}%
}

\begin{document}
\bstctlcite{IEEEexample:BSTcontrol}

\maketitle
\thispagestyle{empty}
\pagestyle{empty}

\begin{abstract}
  World models are useful for robotic manipulation because robots can
  predict how actions change the states of objects before executing them.
  We present \method, a point-set world model that spans rigid, articulated, and
  deformable object manipulation.
  Its state is a set of 3D points on the object and the end-effector, mesh-free
  and topology-agnostic.
  Each point keeps its identity and is supervised on its own trajectory, which
  teaches the model where every point goes rather than only the shape the points
  form.
  Its backbone is a diffusion transformer that denoises a short window of future
  point positions, conditioned on the points' recent history and the commanded
  end-effector motion.
  The backbone's attention alternates between local and global, and
  cross-attention to the end-effector carries the coupling.
  This one architecture at 19.8M parameters and one training recipe cover four
  regimes, rigid objects, cloth, rope, and multi-joint cabinets, with a
  separate checkpoint trained for each.
  Trained on randomized simulation and scored against four baselines on the
  same metric, it is best on three of four regimes and second on rigid.
  Trained on a real-world robot teleoperation dataset, it has the lowest mean
  error in four of its six categories, is second in the other two, and improves
  on the dataset's own model in all six; zero-shot, its simulation checkpoints
  are best on two of four captures.
  Frozen inside sampling-based model-predictive control at one network
  evaluation per window, it plans four simulated tasks over 64 episodes,
  competitive with or outperforming every baseline on each.
  Project website at \url{https://pointcast-wm.github.io}.
\end{abstract}

\section{Introduction}

A robot that pushes a box, drags a cloth, and opens a cabinet is asking the same
question three times: if I move my end-effector like this, what happens next?
A world model answers it, and for manipulation the natural state to predict is
3D geometry.
Dynamics act on shape and contact rather than appearance, and goals are stated
and verified geometrically.
That geometry is also what a simulator labels exactly and generates at no cost,
which lets a geometric world model train on more episodes than a robot can
collect.
Three lines of work have pursued such a model, and each gives up a different
axis.
\emph{Structure-embedded models} build the physics into the
network~\cite{li2018learning,sanchezgonzalez2020gns,zhang2024adaptigraph,zhang2025particle}.
The physics is what makes them accurate inside one regime and what pins them
to it, and where the physics is explicit, the material parameters must be
identified again for every new object.
\emph{Pixel-space world models} buy breadth with
scale~\cite{gao2026dreamdojo,wang2026interactive}, but pixels entangle motion
with appearance and cost more to generate than the geometric state a planner
consumes.
\emph{Point-motion world models} keep the state geometric, and the two closest
to ours still add something to it: appearance features from a frozen image
encoder~\cite{huang2026pointworld} or a material label on every
particle~\cite{huang2025particleformer}.
What remains open is a world model told nothing about an object but where its
points are, and one that covers rigid, articulated, and deformable objects
under one recipe.

\begin{figure}[t]
  \centering
  \includegraphics[width=\columnwidth]{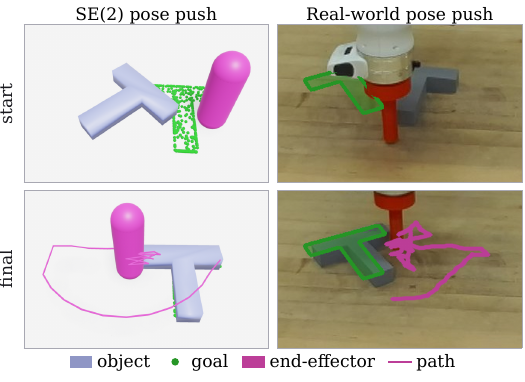}
  \caption{\method{} frozen inside sampling-based MPC, planning from a
  point-set goal.
  One recorded episode each, at the first window the planner controls and
  after the last; the real panel is qualitative.
  Goal green, the executed end-effector path magenta, the pose goal
  flattened to the footprint the block must cover.
  Sec.~\ref{sec:planning} plans three further tasks.}
  \label{fig:teaser}
\end{figure}

\begin{figure*}[!t]
  \centering
  \includegraphics[width=0.94\textwidth]{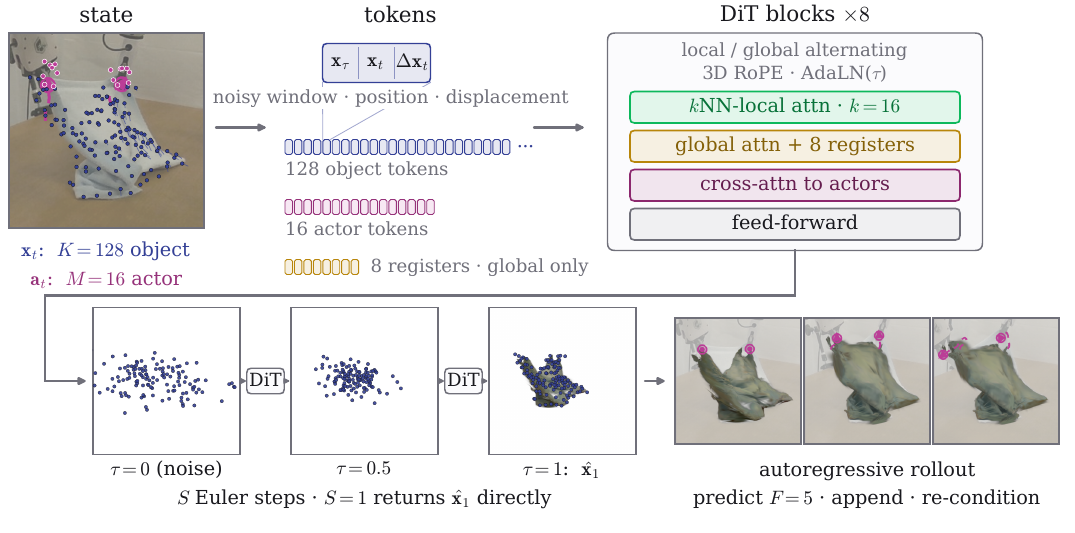}
  \caption{Overview of \method{}.
  Every object point, actor point, and register enters as one token.
  The eight blocks alternate kNN-local and global self-attention, and each
  reads the actor tokens through cross-attention.
  At deployment, one Euler step turns a noise sample into the predicted
  window.}
  \label{fig:arch}
\end{figure*}

We introduce \method, a \emph{point-set world model} whose state is a
persistent set of 3D points on the object and end-effector surfaces
(Fig.~\ref{fig:arch}), with no appearance features, no material label, and no
material parameters.
Each point keeps its identity over time and is supervised on its own
trajectory.
A score that compares shapes without pairing points is met by a cloth in the
right outline with its corners exchanged.
Such a match cannot tell a planner which part moved where.
A diffusion transformer (DiT)~\cite{peebles2023scalable} reads the points and their last displacement,
then denoises the whole next window of point positions at once, with its attention
alternating between a point's neighbors and the whole set.
Cross-attention to the end-effector tokens carries the commanded motion to
every object point.
One recipe trains it on randomized simulation across rigid objects, cloth,
rope, and multi-joint cabinets, or on the PGND benchmark's real robot
episodes~\cite{zhang2025particle}.

We compare against four baselines that also predict a geometric state: the
graph network AdaptiGraph~\cite{zhang2024adaptigraph}, the particle-grid model
PGND~\cite{zhang2025particle}, the capacity-matched point transformer
PTv3~\cite{wu2024ptv3}, and our closest relative,
ParticleFormer~\cite{huang2025particleformer}.
In held-out simulation, \method{} is best on three of four regimes and second
on rigid.
Trained from scratch on the PGND benchmark's own robot episodes, it is best in
four of six categories, second in the other two, and outperforms PGND in all
six (Table~\ref{tab:indomain}, Fig.~\ref{fig:gallery-real}).
Applied zero-shot, its simulation checkpoints are best on two of four
captures (Sec.~\ref{sec:sim2real}).

Our major contributions include:
(i) \method{} is a point-set world model that supervises every point on its own
trajectory, learns how the points move together by attention alternating
between a point's neighbors and the whole set, where any two points can
interact, and predicts how the end-effector moves them by cross-attention.
(ii) Extensive experiments cover four simulated regimes, six real-world
categories, and four zero-shot captures under one architecture, one training
recipe and one checkpoint per regime or category, with four baselines in
simulation and three on real data scored on the same metric.
(iii) Frozen inside sampling-based MPC at one network evaluation per window,
\method{} plans all four tasks (Figs.~\ref{fig:teaser}
and~\ref{fig:planning-compare}), competitive with or outperforming every
baseline on each.

\section{Related Work}
\label{sec:related}

\textbf{Physics-embedded Models and Digital Twins.}
Physics-embedded models couple physics structure with learning, through
particle-grid and material-point
methods~\cite{zhang2025particle,chen2026empm,zhang2024physdreamer},
spring-mass reconstructions~\cite{jiang2025phystwin},
Gaussian representations driven by position-based
dynamics~\cite{abou-chakra2024physically}, and reduced-order elastic
bases~\cite{vismay2024simplicits}.
Digital twins reconstruct a real scene into a simulator that trains or evaluates a policy~\cite{jiang2025gsworld,pfaff2025scalable,jain2025polaris,zhang2025real,cai2026gausstwin,manyar2025realtosim}.
Choosing a solver fixes the equations of motion before any data arrives.
The model is then accurate on the material those equations describe, confined
to it, and dependent on parameters identified for every new object.

\textbf{Learned Point Dynamics.}
Graph-based neural dynamics learn motion from interaction over particles
joined by a hand-set
graph~\cite{li2018learning,sanchezgonzalez2020gns,zhang2024adaptigraph,zhu2024latent}.
They model deformable behavior well and plan through the learned model, with
AdaptiGraph~\cite{zhang2024adaptigraph} estimating physical properties online
to cover unknown materials.
The graph is built from hand-set thresholds on how close two particles must be
to interact, and those thresholds limit which interactions the model can learn.
Point-track and 3D-flow
forecasters~\cite{wen2024atm,bharadhwaj2024track2act,yuan2024generalflow,zhi20253dflowaction}
predict scene-point motion from a goal or an instruction rather than from the
robot's action, and no candidate action can be scored through the model.
Transformer world models drop the graph for point sets or tokens and
condition on the
action~\cite{whitney2024particle,peri2026point,chu2026generative,huang2025particleformer,huang2026pointworld}.
ParticleFormer~\cite{huang2025particleformer}, our closest relative, trains
with Chamfer and Hausdorff losses that compare shapes without pairing points,
and tags every particle with a material label given at test time.
PointWorld~\cite{huang2026pointworld} reaches open-world breadth from 500 hours
of real and simulated video, attaching appearance features from a frozen
DINOv3 encoder~\cite{simeoni2025dinov3} to every point.
\method{} keeps only the points, with no material label and no appearance
features, and attention learns how they move together and how the end-effector
manipulates them, under one recipe for rigid, articulated, and deformable
objects.

\textbf{Video World Models.}
Diffusion policy~\cite{chi2023dp} generates robot actions with a diffusion
transformer, and video world models generate frames with the same
machinery~\cite{lipman2023flow,esser2024sd3,peebles2023scalable},
casting dynamics as conditional video generation.
They are trained to denoise future frames
autoregressively~\cite{chen2024diffusion,huang2026self} and scaled to
manipulation and physics-grounded
video~\cite{gao2026dreamdojo,wang2026interactive,shen2026phantom,wang2026physctrl}.
The pixel objective spends capacity on illumination, texture, and viewpoint.
\method{} keeps the generative formulation on a geometric state and predicts a
whole window in one Euler step, $16.2$\,ms of model compute, and the same model
plans inside sampling-based MPC (Sec.~\ref{sec:planning}).

\section{Method}
\label{sec:method}

\subsection{Problem Setting}
\method{} treats an interaction as a controlled sequence of point sets.
At frame $t$, the object is a set of $K$ control points
$\mathbf{x}_t \in \mathbb{R}^{K\times 3}$.
They are chosen once per interaction from the raw observation by farthest
point sampling~\cite{eldar1994fps}, and row $i$ of $\mathbf{x}_t$ stays at the
same spot on the object at every frame, with the correspondence supplied by
the simulator or, on real data, by a point tracker.
The $M$ actor points $\mathbf{a}_t \in \mathbb{R}^{M\times 3}$ are on the
surface of the end-effector, or of the two grippers in bimanual episodes.
Let $\mathcal{C}_t = (\mathbf{x}_t, \Delta\mathbf{x}_t, \mathbf{a}_{t:t+F})$
collect the current geometry, its one-step displacement
$\Delta\mathbf{x}_t = \mathbf{x}_t - \mathbf{x}_{t-1}$, and the actor
trajectory over the next $F$ frames, which the controller has committed to.
The goal is to fit $\theta$ so that the observed future is as likely as
possible under $p_\theta$,
\begin{equation}
\max_\theta \ \mathbb{E}_{(\mathcal{C}_t,\, \mathbf{x}_{t+1:t+F}) \sim
\mathcal{D}} \big[ \log p_\theta(\mathbf{x}_{t+1:t+F} \mid \mathcal{C}_t)
\big],
\label{eq:problem}
\end{equation}
where $p_\theta$ is the distribution the network places over the next $F$
object states given $\mathcal{C}_t$, and $\mathcal{D}$ is the set of recorded
interactions.
We use $K{=}128$ control points, $M{=}16$ actor points, and a window of
$F{=}5$ frames.
Positions are expressed in the world frame and scaled per interaction by the
object's bounding-box diagonal at the first frame.

\subsection{Architecture}
\begin{figure}[t]
  \centering
  \includegraphics[width=\linewidth]{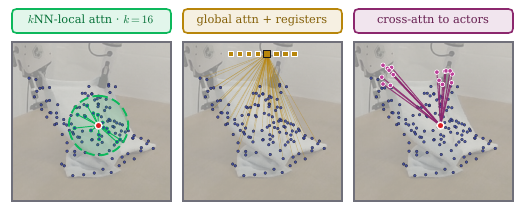}
  \caption{The three attention patterns of the backbone.
  kNN-local restricts a query point (red) to its $k$ nearest neighbors, the
  global panel is drawn from a register token (gold), which attends to every
  object token, and cross-attention reads the actor tokens (magenta).}
  \label{fig:attn}
\end{figure}

The backbone is a diffusion transformer~\cite{peebles2023scalable} over
per-point tokens (Fig.~\ref{fig:arch}).
Each object point contributes one token that fuses its noisy future window
with its observed position and one-step displacement.
That token decodes to the point's $F$ future positions.
All $F$ frames are denoised at once, and attention needs no causal mask in
time.
The flow time of Sec.~\ref{sec:recipe} is not a token but modulates every
block through adaptive layer normalization~\cite{peebles2023scalable}.
In self-attention, 3D rotary position embeddings (RoPE) rotate queries and keys by
the observed positions, and attention depends on the offset between two points.
Three further choices govern how these tokens interact (Fig.~\ref{fig:attn}).
First, self-attention alternates across blocks between a kNN-masked local form
($k{=}16$ neighbors) and a global form, as in the alternating frame-global
attention of large 3D models~\cite{wang2025vggt}.
Second, we add $n{=}8$ register tokens~\cite{darcet2024vision}, learnable
tokens that belong to no point, are excluded from local self-attention, and
are dropped at the output.
Third, each actor point becomes its own cross-attention token carrying its
position and motion at the window's first frame, and each later frame adds one
token holding the end-effector's mean displacement, so finger motion and
end-effector rotation inside the window are not observed.
An object point can then resolve which part of the end-effector is near it
and read contact off the geometry itself.

\begin{figure*}[!b]
  \centering
  \includegraphics[width=\textwidth]{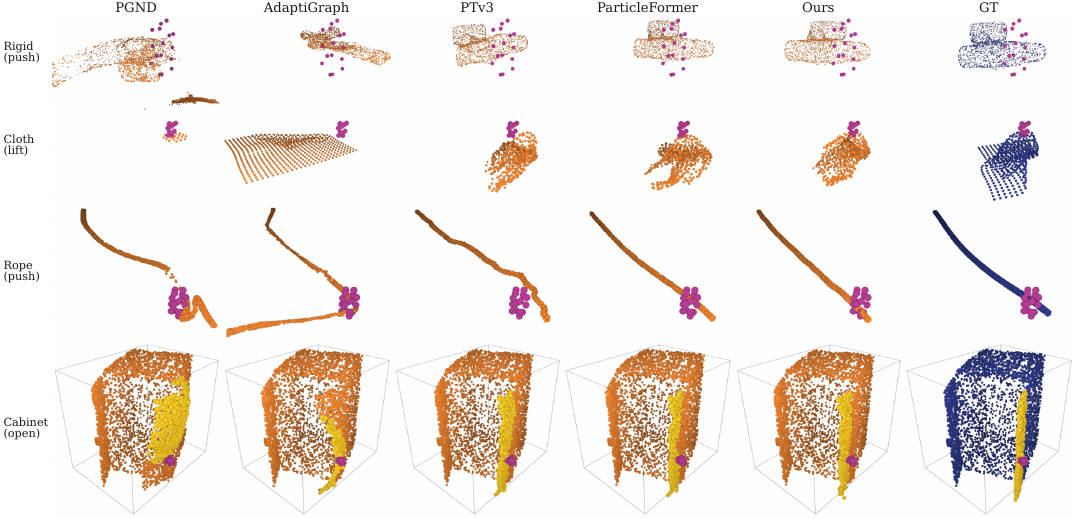}
  \caption{Each model rolled forward from its own predictions through one
  held-out episode per regime, near the end of the interaction.
  Ground truth blue, predictions orange, actor points magenta.
  The cabinet row highlights the moving part in yellow.}
  \label{fig:gallery-sim}
\end{figure*}

\subsection{Objective and Inference}
\label{sec:recipe}
We train Eq.~\eqref{eq:problem} with flow matching~\cite{lipman2023flow}.
Write $\mathbf{z}_\tau$ for the whole window at flow time $\tau$, the linear
interpolation between Gaussian noise $\mathbf{z}_0$ and the clean window
$\mathbf{z}_1$, where the subscript indexes flow time.
We draw $\tau$ from a logit-normal distribution shifted toward the noisy end
of the path~\cite{esser2024sd3}.
The network $f_\theta$ regresses the clean window under three terms,
\begin{equation}
\mathcal{L} = \mathcal{L}_{\text{flow}}
+ \lambda_{\text{v}}\mathcal{L}_{\text{vel}}
+ \lambda_{\text{a}}\mathcal{L}_{\text{dist}} ,
\label{eq:total}
\end{equation}
with $\lambda_{\text{v}}{=}0.1$ and $\lambda_{\text{a}}{=}0.01$ in every
regime.
The first is the squared error of the predicted window, averaged over points
and frames,
\begin{equation}
\mathcal{L}_{\text{flow}} = \mathbb{E}_{\mathbf{z}_1, \tau, \mathbf{z}_0}
\big\| f_\theta(\mathbf{z}_\tau, \tau, \mathcal{C}_t) - \mathbf{z}_1 \big\|^2 .
\label{eq:loss}
\end{equation}
The second matches the velocity the prediction implies to the true path
velocity $\mathbf{z}_1 - \mathbf{z}_0$,
\begin{equation}
\mathcal{L}_{\text{vel}} = \mathbb{E}_{\mathbf{z}_1, \tau, \mathbf{z}_0}
\Big\| \frac{f_\theta(\mathbf{z}_\tau, \tau, \mathcal{C}_t) - \mathbf{z}_\tau}{1-\tau}
- (\mathbf{z}_1 - \mathbf{z}_0) \Big\|^2 .
\label{eq:vel}
\end{equation}
On the linear path Eq.~\eqref{eq:vel} is Eq.~\eqref{eq:loss} reweighted by
$(1-\tau)^{-2}$, and it weights the clean end of the path.
The third is a distance-preservation prior, which expresses what the
per-point terms cannot.
Write $\hat{\mathbf{x}}^i$ for point $i$'s predicted position at one of the
$F$ frames.
The prior holds each point at its initial distance
from its six nearest neighbors~\cite{belyaev2007arap},
\begin{equation}
\mathcal{L}_{\text{dist}} = \frac{1}{W}\sum_i \sum_{j \in \mathcal{N}(i)}
w_{ij}\big(\|\hat{\mathbf{x}}^i - \hat{\mathbf{x}}^j\| - d_{ij}\big)^2 ,
\label{eq:arap}
\end{equation}
where $\mathcal{N}(i)$ holds point $i$'s six nearest neighbors, $d_{ij}$ is
the rest length, and $w_{ij} = e^{-\gamma d_{ij}^2}$ is its weight, all read off
the first frame, and $W$ sums the weights.
The term is summed over the $F$ predicted frames and is identical in every
regime except for how the neighbors are chosen.
On the cabinet, a point's neighbors come only from points on the same part,
which is the only place a part label enters the loss.
ParticleFormer trains on a shape term~\cite{huang2025particleformer}
\emph{instead of} Eq.~\eqref{eq:loss}, an equal mix of the symmetric Chamfer
and Hausdorff distances that compares shapes without pairing points.

At inference, we draw $\mathbf{z}_0$ from a standard normal distribution and
integrate the flow from $\tau{=}0$ to $\tau{=}1$ in $S$ Euler steps, and we
deploy $S{=}1$, where the single step returns the predicted clean window
$\hat{\mathbf{x}}_1$ directly.
A rollout slides the window forward, appending the $F$ predicted frames to
the history and reading the next stretch of actor trajectory from the
controller.

\section{Experiments}
\label{sec:experiments}

\subsection{Setup}
\begin{figure*}[t]
  \centering
  \includegraphics[width=\textwidth]{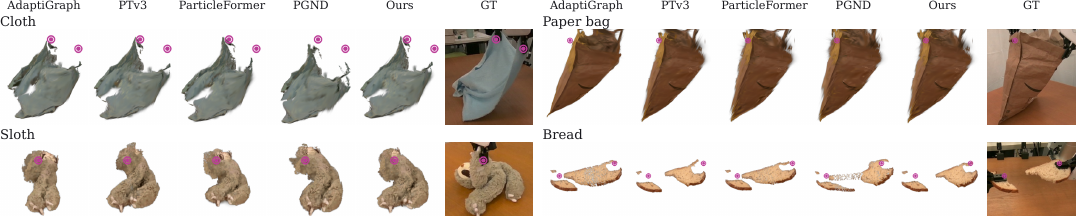}
  \caption{Real-world rollouts on the PGND benchmark at step 30.
    GT is the photograph, and each prediction column shows the episode's
    pre-interaction Gaussian splats carried by that method's predicted point
    motion~\cite{zhang2025particle}.
  AdaptiGraph is shown but not scored (Sec.~\ref{sec:sim2real}).}
  \label{fig:gallery-real}
\end{figure*}

\textbf{Data.}
The simulated data come from the Newton simulator~\cite{newton2026}.
The end-effector pushes rigid YCB objects, grasps and lifts cloth, pushes and
grasps rope, and grasps the handle of cabinets with revolute and prismatic
joints.
Object geometry, material parameters, and commanded motion are drawn at random
per episode.
The real data are the six-category PGND benchmark~\cite{zhang2025particle},
multi-view RGB-D recordings of teleoperated robot interactions, used under
its own splits and metric.
Object points come from the simulator's particle set and, on real data, from
the benchmark's CoTracker3~\cite{karaev2025cotracker} tracks as released.
Actor points are the same gripper-finger template in both, posed on real data
along the benchmark's commanded end-effector trajectory; the template carries
no gripper-open distance, and on bimanual captures the $16$ points split
across the two grippers.
In simulation, data are split by object, and every evaluation object is unseen
in training.
Training pools hold 112k rigid, 10.2k cloth, 10.2k rope, and 5.0k cabinet
interactions, with $15\%$ held out for validation.

\textbf{Models.}
AdaptiGraph~\cite{zhang2024adaptigraph} and PGND~\cite{zhang2025particle} keep
their authors' recipes and harnesses, with PGND running from its released
checkpoints on the real benchmark, marked native in Table~\ref{tab:indomain}.
Since ParticleFormer~\cite{huang2025particleformer} released neither code nor
hyperparameters, we reimplemented it from the paper and give its full
specification on the project website.
PTv3~\cite{wu2024ptv3} is the backbone PointWorld~\cite{huang2026pointworld}
uses, driven by our objective and one-step sampler, with the actor points
appended to its point set and every object point given its offset to the
nearest actor point.
PTv3, ParticleFormer, and \method{} train from scratch per regime and
category on the same cosine-decay schedule, one checkpoint each, selected on
a validation slice and scored once.
Each trains for $200$ epochs of $800$ steps at batch size $64$ on the same
per-regime budget, about $7$ hours on one A100 or H100.
A training window, the $F{=}5$ future frames of Sec.~\ref{sec:method}, takes
consecutive frames for rigid and cabinet and every fourth frame for cloth and
rope.

\textbf{Evaluation.}
In simulation, a rollout follows the ground-truth end-effector trajectory
through the interaction and carries the dense cloud by linear-blend skinning
from the predicted control points.
The metric is the mean rollout error over that cloud (mm) across every frame
of every interaction; on the cabinet it scores the actuated joint only.
An interaction holds $86$ frames at $15$\,Hz for rigid, $114$ for cloth, $34$
to $75$ for rope, and $40$ for the cabinet.
On real data, every method is scored under the benchmark's own protocol: the
mean displacement error at the 3\,s horizon of Table~\ref{tab:indomain}.
AdaptiGraph and PGND sample their own control points in simulation rather than
ours.

\subsection{Fidelity in Simulation and on Real Data}
\label{sec:indomain}

\subsubsection{Simulation}
Table~\ref{tab:sim} shows the mean rollout error in the simulation benchmark.
\method{} is best on three of the four regimes.
It leads ParticleFormer by $16.6\%$ on cloth and $32.6\%$ on rope.
On the cabinet actuated joint, it leads PTv3, $6.84$ against $10.34$.
Rigid is the one loss, $16.9\%$ behind ParticleFormer
(Fig.~\ref{fig:gallery-sim}).
Sec.~\ref{sec:ablations} traces the gap to the per-point loss rather than the
backbone.
PTv3 loses every regime at matched budget and capacity (19.5M parameters
against our 19.8M), placing the gain in the architecture.

\begin{table}[t]
\caption{Simulation benchmark, mean rollout error (mm)}
\label{tab:sim}
{\centering
\begin{tabular}{lcccc}
\toprule
Method & Rigid & Cloth & Rope & Cabinet$^{\ast}$ \\
\midrule
PGND~\cite{zhang2025particle} & 47.82 & 524.33 & 68.76 & 54.45 \\
AdaptiGraph~\cite{zhang2024adaptigraph} & 55.42 & 153.14 & 49.24 & 36.64 \\
PTv3~\cite{wu2024ptv3} & 52.06 & 42.82 & 72.30 & \underline{10.34} \\
ParticleFormer~\cite{huang2025particleformer} & \textbf{20.88} & \underline{38.26} & \underline{21.78} & 26.73 \\
\midrule
\rowcolor{gray!15}
\method{} (ours) & \underline{24.40} & \textbf{31.90} & \textbf{14.67} & \textbf{6.84} \\
$n$ (interactions) & 2796 & 1047 & 1119 & 996 \\
\bottomrule
\end{tabular}
\par}
\vspace{2pt}
{\footnotesize $^{\ast}$actuated joint. \textbf{Best}, \underline{second best}.}
\end{table}

\begin{table}[t]
\caption{Real-world PGND benchmark, mean displacement error at the
3\,s horizon (mm)}
\label{tab:indomain}
{\centering
\setlength{\tabcolsep}{3.4pt}
\begin{tabular}{lcccccc}
\toprule
Method & Cloth & Rope & Box & Bread & Bag & Sloth \\
\midrule
PTv3~\cite{wu2024ptv3} & 42.24 & 25.85 & \textbf{21.76} & 22.37 & 21.40 & 48.43 \\
ParticleFormer~\cite{huang2025particleformer} & \textbf{31.65} & \underline{24.10} & 24.54 & 22.61 & 18.62 & 44.32 \\
PGND~\cite{zhang2025particle} (native) & 44.49 & 39.14 & 27.79 & \underline{20.33} & \underline{16.16} & \underline{42.94} \\
\midrule
\rowcolor{gray!15}
\method{} (ours) & \underline{32.20} & \textbf{15.83} & \underline{23.51} & \textbf{18.65} & \textbf{11.64} & \textbf{37.83} \\
$n$ (test episodes) & 40 & 40 & 20 & 20 & 20 & 20 \\
\bottomrule
\end{tabular}
\par}
\vspace{2pt}
{\footnotesize \textbf{Best}, \underline{second best}.
The benchmark's released evaluation scores step 30, the end of its 3\,s
horizon, and reproduces the published values within $1$\,mm on five of six
categories.}
\end{table}

\subsubsection{Real World}
\label{sec:sim2real}
On real data, we run one evaluation in-domain on the benchmark's own episodes
and one transfer from the simulation checkpoints.
AdaptiGraph is scored in neither, because its released evaluation advances only
while the end-effector moves and teleoperated captures pause.

\textbf{In-domain.}
\method{} has the lowest mean error in four of the six categories of
Table~\ref{tab:indomain} and is second in the other two
(Fig.~\ref{fig:gallery-real}).
It is the only method that improves on PGND in all six.

\begin{figure*}[t]
  \centering
  \includegraphics[width=\textwidth]{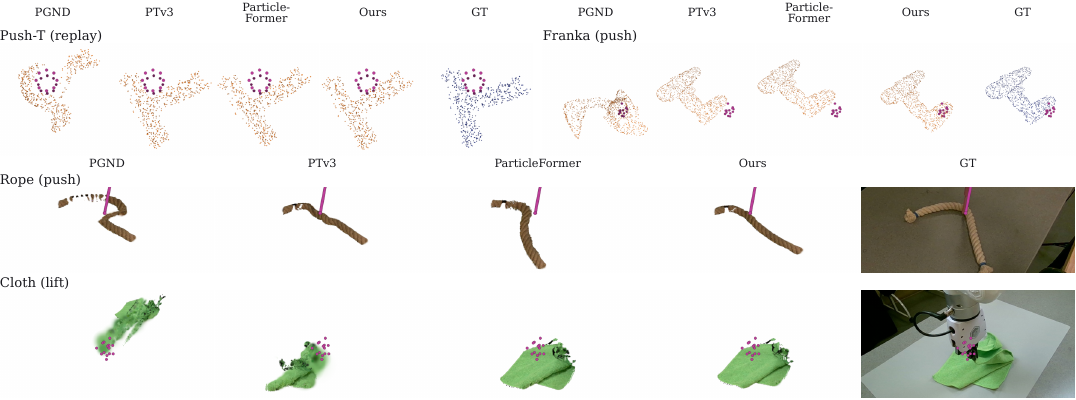}
  \caption{Zero-shot rollouts, one episode per capture of
    Table~\ref{tab:real}.
    GT is a point cloud on Push-T and YCB objects and the photograph on rope and
    cloth, and the end-effector drawn is the recorded one, not the actor points
  the models read.}
  \label{fig:gallery-zeroshot}
\end{figure*}

\begin{table}[t]
\caption{Zero-shot transfer, rollout error (mm)}
\label{tab:real}
{\centering
\setlength{\tabcolsep}{4pt}
\begin{tabular}{lcccc}
\toprule
Method & Push-T replay & Franka YCB & Rope & Cloth \\
\midrule
PGND~\cite{zhang2025particle} & 37.27 & 75.38 & 58.07 & 291.12 \\
PTv3~\cite{wu2024ptv3} & 24.67 & \underline{60.30} & 83.45 & 54.77 \\
ParticleFormer~\cite{huang2025particleformer} & \underline{18.07} & 72.50 & \textbf{42.47} & \textbf{32.43} \\
\midrule
\rowcolor{gray!15}
\method{} (ours) & \textbf{15.35} & \textbf{52.86} & \underline{49.05} & \underline{42.20} \\
$n$ (interactions) & 70 & 35 & 123 & 31 \\
\bottomrule
\end{tabular}
\par}
\vspace{2pt}
{\footnotesize \textbf{Best}, \underline{second best}.
Push-T and Franka pool the frames with end-effector contact, rope and cloth
average per interaction.
Baselines run at their frozen settings, \method{} at its
deployed one-step sampler.}
\end{table}

\textbf{Zero-shot transfer.}
The simulation checkpoints of Table~\ref{tab:sim} are applied without any real
data to four captures whose motions match the training actions, Push-T
simulator replay~\cite{chi2023dp}, rope pushing and cloth grasp
lifts~\cite{zhang2024dynamic}, and our own Franka Emika Panda pushes of
YCB objects held out of the simulation pool.
On rope, cloth, and the Franka pushes, ground truth comes from
multi-view reconstruction and
tracking~\cite{luiten2024dynamic,karaev2025cotracker}, with
FoundationPose~\cite{wen2024foundation} on the YCB objects.
\method{} is the best of the four methods on Push-T and Franka, and
ParticleFormer leads rope and cloth (Table~\ref{tab:real},
Fig.~\ref{fig:gallery-zeroshot}).
The same cloth and rope checkpoints score $92.45$ and $134.33$\,mm on the PGND
episodes, whose end-effector barely moves per frame; re-striding recovers
much of that, $65.25$ and $51.87$\,mm, leaving a gap of $2.0\times$ on cloth
and $3.3\times$ on rope over the in-domain rows.

\begin{table}[t]
\caption{Simulation ablations, percent change from the shaded row}
\label{tab:ablation}
{\centering\small
\setlength{\tabcolsep}{2.0pt}
\begin{tabular}{lcccc}
\toprule
Variant & Rigid & Cloth & Rope & Cabinet$^{\ast}$ \\
\midrule
\rowcolor{gray!15}
\method{} (Table~\ref{tab:sim}, flow, $S{=}1$) & 24.40 & 31.90 & 14.67 & 6.84 \\
\multicolumn{5}{l}{\emph{What decides articulation}} \\
\;\; ParticleFormer set-level loss & $-14.5$ & $+3.5$ & $-2.2$ & $+79.3$ \\
\;\; no distance prior & $+1.7$ & $-3.1$ & $-2.1$ & $+18.0$ \\
\;\; part-aware prior off (cabinet) & -- & -- & -- & $+15.6$ \\
\multicolumn{5}{l}{\emph{Attention and tokenization}} \\
\;\; local-only attention$^{\dagger}$ & $+4.9$ & $-1.5$ & $-2.1$ & $+6.9$ \\
\;\; global-only attention$^{\dagger}$ & $+8.8$ & $-2.3$ & $-1.7$ & $-5.4$ \\
\;\; block-causal, per-frame tokens & $+33.1$ & $+124$ & $+248$ & $+147$ \\
\multicolumn{5}{l}{\emph{Sampler and checkpoint}} \\
\;\; $S{=}25$ & $+5.3$ & $-0.9$ & $+0.4$ & $+16.1$ \\
\;\; direct regression & $-5.1$ & $-2.3$ & $+2.0$ & $+4.1$ \\
\;\; joint checkpoint, $4\times$ budget & $+3.3$ & $-0.1$ & $+15.4$ & $+284$ \\
\bottomrule
\end{tabular}
\par}
\vspace{2pt}
{\footnotesize Shaded: the model of Table~\ref{tab:sim}, in mm.
Every other row is percent change against it, positive worse,
$^{\dagger}$against the earlier recipe's own base.
$^{\ast}$actuated joint.}
\end{table}

\begin{table*}[b]
  \centering
  \caption{Planning across four tasks}
  \label{tab:planning}
  {\small
    \setlength{\tabcolsep}{3pt}
    \begin{tabular}{llccccc>{\columncolor{gray!15}}c}
      \toprule
      Task & Metric & init & AdaptiGraph & PGND & PTv3 & ParticleFormer & \method{} (ours) \\
      \midrule
      SE(2) pose push & per-point (cm) & --   & 9.52\,{\scriptsize(0.19)} & 2.31\,{\scriptsize(5.98)} & 2.27\,{\scriptsize(0.25)} & \underline{1.57}\,{\scriptsize(0.17)} & \textbf{1.39}\,{\scriptsize(0.10)} \\
      Sequential articulation & part err (cm) & -- & 7.97 / 2.67\,{\scriptsize(0.07)} & 2.25 / \textbf{0.38}\,{\scriptsize(1.90)} & \underline{0.72} / 0.97\,{\scriptsize(0.21)} & 0.88 / \underline{0.41}\,{\scriptsize(0.15)} & \textbf{0.42} / \textbf{0.38}\,{\scriptsize(0.09)} \\
      Cloth drag      & Chamfer (cm)   & 27.8 & 27.75\,{\scriptsize(0.27)} & 2.83\,{\scriptsize(2.23)} & 2.41\,{\scriptsize(0.03)} & \textbf{2.08}\,{\scriptsize(0.03)} & \underline{2.09}\,{\scriptsize(0.04)} \\
      Rope drag       & Chamfer (cm)   & 15.4 & 3.28\,{\scriptsize(0.31)} & \underline{1.91}\,{\scriptsize(12.08)} & 5.87\,{\scriptsize(0.03)} & 2.16\,{\scriptsize(0.03)} & \textbf{1.71}\,{\scriptsize(0.04)} \\
      \bottomrule
    \end{tabular}
  \par}
  \vspace{2pt}
  \begin{minipage}{\textwidth}
    {\footnotesize Mean distance error over 16 episodes, and in parentheses
      the planning time per window, in seconds on one RTX 4090 with the
      rollout batched over the task's $N$ candidates. \emph{init} is the
      starting distance, best \textbf{bold}, second \underline{underlined};
    the sequential cells give the first and the second commanded joint.}
  \end{minipage}
\end{table*}

\begin{figure*}[t]
  \centering
  \includegraphics[width=\textwidth]{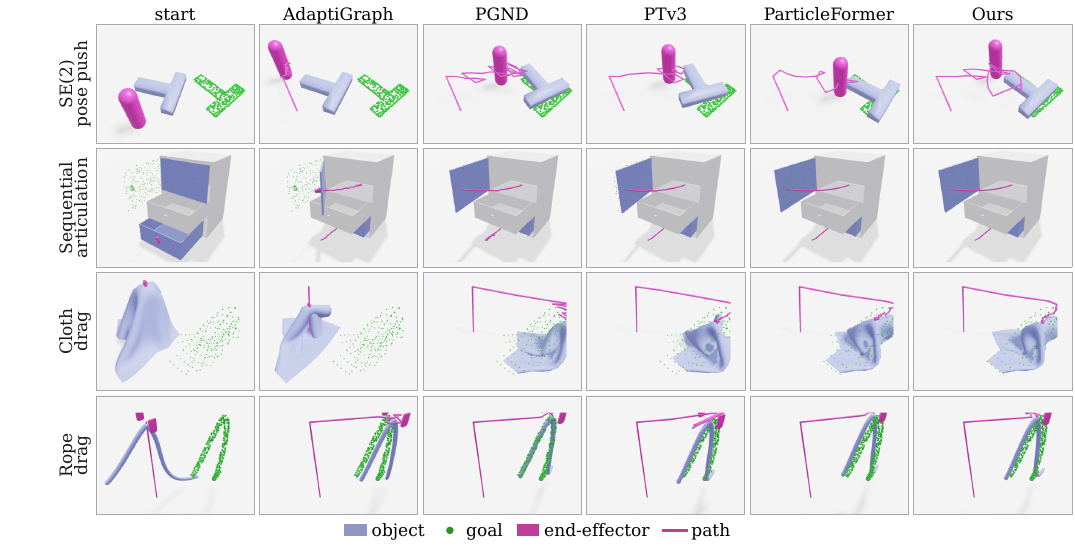}
  \caption{One recorded episode per task, the start and the final state each
    rollout model leaves under the same planner, cost, budget, seed, and goal.
    Object blue, goal green, end-effector magenta with its executed path, the
    pose goal flattened to the footprint the block must cover.
  Table~\ref{tab:planning} carries the statistics.}
  \label{fig:planning-compare}
\end{figure*}

\subsection{Ablations}
\label{sec:ablations}

Table~\ref{tab:ablation} varies one piece of the model at a time, and its last
block replaces the recipe instead.
Every row keeps the training seed and the one-step scoring, apart from the
$S{=}25$ and direct-regression rows.

\textbf{Per-point supervision and the distance prior decide articulation.}
ParticleFormer's set-level objective of Sec.~\ref{sec:recipe} scores the shape
the points form instead of where each point went.
The swap makes the cabinet's actuated joint $79\%$ worse, while the error over
all cabinet points improves, $9.66$ to $6.51$.
A shape match is satisfied by the static body and gives up the moving part.
The per-point loss therefore buys the moving part at the cost of the rigid
regime.
Cloth gets $3.5\%$ worse as well, and rope improves by $2.2\%$.
Rigid is the regime the swap helps most, by $14.5\%$, and there it lands our
backbone at ParticleFormer's own rigid number ($20.86$ against $20.88$).
Under either loss, the model beats ParticleFormer on cloth, rope, and the
cabinet.
The distance prior earns its place on the same joint: removing it costs the
cabinet joint $18\%$, improves cloth by $3.1\%$ and rope by $2.1\%$, and
moves rigid by $1.7\%$.
The one cabinet-specific setting in the recipe is the part label picking that
prior's neighbors, worth $15.6\%$ on the actuated joint.

\textbf{The alternating pattern and the whole-window tokens earn their place.}
Both single-pattern arms are worse than the alternation on rigid, by $4.9\%$
and $8.8\%$, and change cloth, rope, and the cabinet joint by less than $7\%$
in either direction.
The video recipe~\cite{chen2024diffusion} gives every point one token per frame
and attends block-causally across frames with temporal rotary embeddings.
It is worse in every regime, from $33\%$ on rigid to $3.5$ times on rope, and
carries $21\%$ more parameters.

\textbf{One network evaluation is enough, and one checkpoint per regime is not.}
The noise draw moves the output by only $2$ to $17\%$ of the model's own error
on all ten evaluation pools.
Twenty-five steps buy under $1\%$ on cloth and nothing on rope, while making
rigid $5\%$ and the cabinet joint $16\%$ worse.
A direct regression of the same network under the same per-point loss lands
within $5.1\%$ on every regime, placing the gain in the state and the
supervision the two share.
Flow matching stays because one Euler step costs the same and keeps the
training on the generative objective of Eq.~\eqref{eq:problem}.
One checkpoint over all four regimes at four times the budget stays within
$3.3\%$ on rigid and cloth, is $15\%$ worse on rope and $3.8$ times worse on
the cabinet joint.
At deployment, one predicted window takes $16.2$\,ms of model compute on one
RTX 4090 at batch size one, against $384.8$\,ms for $25$ steps ($23.8\times$).
Free-running rollouts run $18$ to $77\times$ faster than real time at the
$15$\,Hz observation rate.

\subsection{Planning with World Model}
\label{sec:planning}

One network evaluation per window (Sec.~\ref{sec:ablations}) makes the frozen
\method{} model cheap enough to drive sampling-based MPC.
At every replanning step, the planner draws $N$ candidate action sequences from
an isotropic Gaussian around a proposal mean and rolls each through the model
in one step from one shared noise draw.
The predicted points are scored against a point-set goal, and the MPPI
rule~\cite{williams2018mppi} pulls the mean toward the lowest-cost candidates
at AdaptiGraph's temperature~\cite{zhang2024adaptigraph}, with the sequential
task refitting to its best candidates.
The planner executes the best candidate for one window before replanning.
We use $N{=}32$, $16$, and $8$ candidates on the push, cabinet, and drag tasks, respectively.
It looks three windows ahead on the pose push, where the end-effector cannot
reposition to a better contact face inside one window, two on the sequential
task, and one on the drags, the same setting for every model.

\textbf{Tasks.}
All four tasks use unseen object geometry, and every goal is a \emph{point-set
configuration} (Fig.~\ref{fig:planning-compare}).
\emph{SE(2) pose push} sends a T-block to a planar target $10$--$20$\,cm away
at a commanded yaw of $20$ to $60$ degrees.
\emph{Sequential articulation} commands two of a cabinet's joints in a sampled
order.
The planner opens the first, releases it, and works the second from what the
first left behind, with the transit between handles scripted for every model.
\emph{Cloth drag} and \emph{rope drag} start from a scripted grasp and move the
object to a translated target.

\begin{figure}[t]
  \centering
  \includegraphics[width=\linewidth]{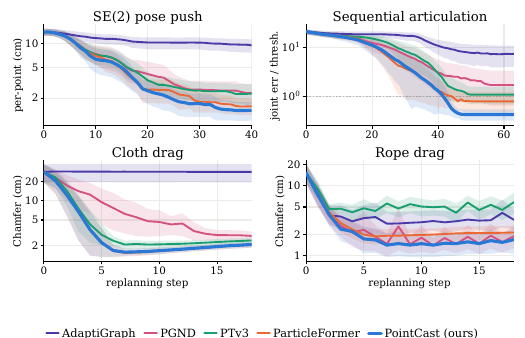}
  \caption{Goal distance per replanning step, log axes, band a bootstrap
    $95\%$ interval over the 16 episodes.}
  \label{fig:planning-curves}
\end{figure}

\textbf{Results.}
Our model plans all four tasks (Table~\ref{tab:planning}), competitive with or
outperforming every baseline on each.
It outperforms PTv3 and AdaptiGraph on pose push and both drags, PGND on pose
push and cloth drag, and ParticleFormer on rope drag.
The remaining margins are within the run-to-run spread.
\method{} has the lowest error on the first commanded joint, ties PGND on the
second, and its curve settles lowest after the re-attach
(Fig.~\ref{fig:planning-curves}).
PGND, one of the weakest free-running predictors in Table~\ref{tab:sim},
recovers once the planner replans every window, but lets the first joint drift
to $2.25$\,cm after the release.
The same planner and the simulation rigid checkpoint, with no real data, push
a printed T-block toward its pose goal on a Franka Emika Panda
(Fig.~\ref{fig:teaser}).
Its points are the printed shape fitted in SE(2) to the fused depth cloud.

\section{Conclusion}
\label{sec:conclusion}

\method{} treats manipulation dynamics as the motion of a persistent point
set, supervised per point.
Under one recipe and one metric, it is best on three of four simulated regimes
and second on rigid.
On a real-robot benchmark, it is best in four of six categories and second in
the other two.
With one Euler step of the sampler predicting a whole window, \method{} plans
four simulated tasks inside sampling-based MPC, competitive with or
outperforming every baseline on each.
Geometry alone, with no material specified by hand, predicts rigid,
articulated, and deformable motion.
\method{} still predicts one window at a time, and a rollout that feeds its own
predictions back compounds their error over the horizon.
The pose push already plans three windows ahead, and going further is a
question of how long a rollout holds.
Video models meet the same limit and answer it with self-forcing and diffusion
forcing~\cite{huang2026self,chen2024diffusion}, and a point set is a state that
can carry the same treatment.

\bibliographystyle{IEEEtran}
\bibliography{refs}

\end{document}